%% file: irt2.tex
\pdfoutput=1

\documentclass[english]{lni}
\usepackage{graphicx}

\usepackage{tikz}
\usepackage{tikz-network}
\usetikzlibrary{arrows.meta}

\usepackage{mathtools}
\usepackage{amsfonts, amsmath, amssymb}
\usepackage{relsize}  
\usepackage{accents}

\usepackage{booktabs, multirow} 
\usepackage{soul}
\usepackage{adjustbox}

\usepackage[nottoc,numbib]{tocbibind}
\usepackage[xindy, toc]{glossaries}
\loadglsentries{glossary}
\glsdisablehyper  

\definecolor{c-blue-1}{RGB}{11, 44, 64}
\definecolor{c-blue-2}{RGB}{21, 88, 127}
\definecolor{c-blue-3}{RGB}{43, 177, 255}
\definecolor{c-blue-4}{RGB}{170, 222, 255}
\definecolor{c-blue-5}{RGB}{224, 243, 255}

\definecolor{c-gold-1}{RGB}{92, 73, 56}
\definecolor{c-gold-2}{RGB}{181, 157, 121}
\definecolor{c-gold-3}{RGB}{255, 221, 171}
\definecolor{c-gold-4}{RGB}{255, 235, 199}
\definecolor{c-gold-5}{RGB}{255, 243, 220}

\definecolor{c-magenta-1}{RGB}{82, 20, 36}
\definecolor{c-magenta-2}{RGB}{107, 64, 76}
\definecolor{c-magenta-3}{RGB}{255, 62, 114}
\definecolor{c-magenta-4}{RGB}{255, 157, 182}
\definecolor{c-magenta-5}{RGB}{255, 216, 232}

\definecolor{c-grey-1}{RGB}{51, 51, 51}
\definecolor{c-grey-2}{RGB}{102, 102, 102}
\definecolor{c-grey-3}{RGB}{153, 153, 153}
\definecolor{c-grey-4}{RGB}{187, 187, 187}
\definecolor{c-grey-5}{RGB}{238, 238, 238}

\definecolor{c-amber-3}{RGB}{255, 145, 0}

\newcommand{\ktzrelation}[1]{{\color{c-grey-2} \text{#1}}}
\newcommand{\ktzentity}[1]{{\color{c-grey-2} \textsc{#1}}}
\newcommand{\ktzmention}[1]{{\color{c-grey-2} \textit{``\underline{#1}''}}}
\newcommand{\ktzmentionI}[1]{{\color{c-grey-2} \textit{\underline{#1}}}}
\newcommand{\ktzsequence}[1]{{\color{c-grey-2} \textit{``#1''}}}

\begin{document}
\title[IRT2 - F. Hamann et al.]{IRT2: Inductive Linking and Ranking in Knowledge Graphs of Varying Scale}
\author[Felix Hamann \and Adrian Ulges \and Maurice Falk]
{
Felix Hamann\footnote{RheinMain University of Applied Sciences, Wiesbaden, Germany \\\email{{felix.hamann, adrian.ulges, maurice.falk}@hs-rm.de}} \and
Adrian Ulges\footnotemark[1] \and 
Maurice Falk\footnotemark[1]
}
\startpage{1} 
\editor{Hamann et al.} 
\booktitle{W6: Text Mining and Generation (TMG-2022)} 
\yearofpublication{2017}
\maketitle

\begin{abstract}
We address the challenge of building domain-specific knowledge models for industrial use cases, where labelled data and taxonomic information is initially scarce. Our focus is on inductive link prediction models as a basis for practical tools that support knowledge engineers with exploring text collections and discovering and linking new (so-called open-world) entities to the knowledge graph. We argue that -- though neural approaches to text mining have yielded impressive results in the past years -- current benchmarks do not reflect the typical challenges encountered in the industrial wild properly. Therefore, our first contribution is an open benchmark coined IRT2 (inductive reasoning with text) that (1) covers knowledge graphs of varying sizes (including very small ones), (2) comes with incidental, low-quality text mentions, and (3) includes not only triple completion but also ranking, which is relevant for supporting experts with discovery tasks.

We investigate two neural models for inductive link prediction, one based on end-to-end learning and one that learns from the knowledge graph and text data in separate steps. These models compete with a strong bag-of-words baseline. The results show a significant advance in performance for the neural approaches as soon as the available graph data decreases for linking. For ranking, the results are promising, and the neural approaches outperform the sparse retriever by a wide margin.

\end{abstract}


\section{Introduction}

Knowledge graphs (KG) are known to be powerful tools in various applications such as web search or personal assistants. They are key to addressing complex information needs that require reasoning and cannot be answered with simple text matching. Usually, knowledge graphs are assumed to be available at large scale~\cite{hogan2021knowledge}, examples include the Linked Open Data Cloud\footnote{https://lod-cloud.net} or KGs in biomedicine~\cite{hoehndorf15biomedical}. However, knowledge graphs may also be interesting in other industries where no large-scale knowledge exists yet, be it in the technical service, where KGs can help reason about the root causes of machine failures, in insurance, where they can model the details of a contract, or in finance rating, where they capture businesses' supply chains. Explicit knowledge is scarce in these domains.  However, what is often commonly available in abundance is text data, such as service tickets, insurance claims, or business reports. 

The focus of this paper is to support experts with building knowledge graphs by inspecting text. To do so, we assume that the expert adds triples to the graph in an iterative process, in which he/she locates and identifies new entities in the text collection and links them to the graph. We address two key steps of this process, as illustrated in Figure \ref{fig:approach}:

\begin{figure}[t]
    \centering
    \includegraphics[width=1\textwidth]{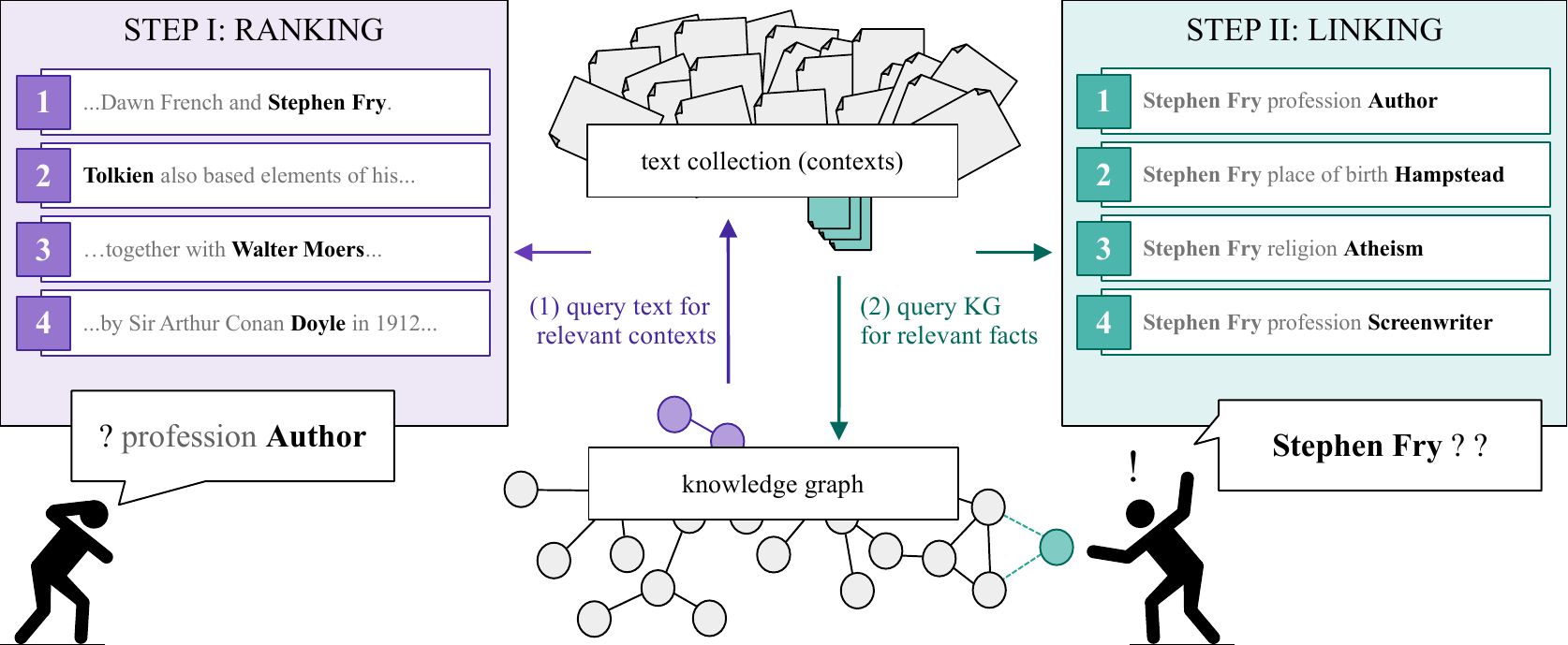}
    \caption{Our two-step approach towards interactive knowledge acquisition: First, the user discovers sentences containing relevant entities (Discovery/Ranking, left). Second, given a mention of interest, triples for this mention are suggested as if it was an entity (Linking, right). }
    \label{fig:approach}
\end{figure}

\begin{enumerate}
    \item {\bf Discovery/Ranking}: The expert explores the text collection based on a partial fact of the KG (i.e. a relation-vertex tuple). He/she specifies an information need (such as ``find me actors'', or more specifically ``find sentences likely to contain mentions of entities which have the relation \ktzrelation{profession} with the known entity \ktzentity{actor}''). The system returns a ranked list of potentially interesting sentences, giving rise to a ranking task.
    \item {\bf Linking}: The expert studies the sentences. Once he/she has discovered an interesting entity $x$ (such as \ktzentity{Frances McDormand}), we would like to link $x$ to the graph by adding triples, not only via the \ktzrelation{profession} relation but also via other relations (such as the birthplace, languages she speaks, or movies she appeared in). To do so, the system collects textual evidence on the entity in form of other mentions in the text collection. From those, it infers triples that link $x$ to the knowledge graph.
\end{enumerate}
We address both tasks using embedding-based \textit{semi-inductive}~\cite{ali2021improving} \textit{open-world}~\cite{shi2017open} link prediction, which is targeted at predicting the likelihood of triples $(h,r,t)$. In contrast to standard link prediction, semi-inductive link prediction addresses {\it new} entities to be linked to the known graph and the open-world scenario connotes that the new entities are solely described via free text.

Since domain-specific data is often confidential, link prediction research employs open data from graphs such as Freebase~\cite{Bollacker:2008:FCC:1376616.1376746} or Wikidata~\cite{vrandevcic2014wikidata}. We argue that the insight these benchmarks offer for industrial knowledge acquisition is limited, and contribute a new benchmark called \textbf{Inductive Reasoning with Text V2 (IRT2)} with the following benefits\footnote{The data is publicly available under \url{https://github.com/lavis-nlp/irt2}}: (1) To assess models at different stages of graph construction, we benchmark on graphs of varying size. (2) While text contexts in other datasets consist of concise descriptions of entities, text in practice contains rather incidental mentions of entities. We sample our text data accordingly. (3) To our knowledge, our study is the first to not only address the linking task but also the ranking task.

We evaluate three inductive link models on IRT2: One baseline based on keyword matching, and two neural models that combine a transformer text encoder with a link predictor. We show that the end-to-end approach and separate training both work well, while the latter generally performs a little better (even for small graphs). Also, only through evaluation on smaller graphs, the baseline model is outperformed when linking, which favours the neural models in scenarios where structured data becomes sparse.

\section{Related Work}

\textbf{Link Prediction Models:} Closed-world link prediction (or also {\it knowledge graph completion, KGC}) has attracted interest in the research community over the past years. Earlier approaches such as RESCAL~\cite{nickel2011three},  TransE~\cite{bordes2013translating}, and DistMult~\cite{yang2014embedding} laid the foundations for many following completion models such as ComplEx~\cite{trouillon2016complex}, ConvE~\cite{dettmers2018convolutional}, RotatE~\cite{sun2019rotate}, or KBGAT~\cite{nathani2019learning}. 
 We are, however, interested in open-world scenarios, where, given a textual description of an entity, geometric reasoning is applied through the alignment of dense graph- and text-representations. This combines KGC---which aims at the prediction of missing links between existing entities---with language modelling~\cite{mikolov2013distributed,bojanowski2016enriching,devlin2018bert} which produces dense vector representations for natural language text. Specifically, in this work, a semi-inductive setting~\cite{ali2021improving} is studied, where out-of-kg entities are to be linked to an existing graph using their textual description. One of the first approaches to combining language- and graph-reasoning models is NTN~\cite{socher2013reasoning}, where entity description embeddings are trained to be similar if their entities are connected in the KG. Later approaches build on this idea: \acrshort{dkrl}~\cite{xie2016representation}, \gls{conmask}~\cite{shi2017open}, \acrshort{mia}~\cite{fu2020multiple}, and KEPLER~\cite{wang2021kepler} introduce models that jointly learn the connection between entity and graph representations. A different approach decouples the KGC scorer and text encoder such that first the scorer is trained independently and in a separate step a projection from text-based entity descriptions to their associated graph-based embeddings is learned~\cite{shah2019open,shah-etal-2020-relation,zhou2020weighted}. We study and compare both approaches with the models described in \Cref{sec:models}. Highly related to our work is the recently published approach by Daza et al~\cite{daza2021inductive}. Here, a BERT model (BLP) is trained to directly score KGC triples for plausibility using the textual representations instead of training a separate entity embedding. However, we cannot rely on high quality text and thus a separate entity embedding tuned with many different observations for input text is more suitable.

\textbf{KGC Benchmarks:} Closed-world KGC models are usually evaluated on subsets of publicly available knowledge graphs such as Freebase~\cite{Bollacker:2008:FCC:1376616.1376746}, DBPedia~\cite{auer2007dbpedia} or Wikidata~\cite{vrandevcic2014wikidata}. Famous benchmarks include FB15K~\cite{bordes2013translating}, succeeded by FB15k237~\cite{toutanova-chen-2015-observed} due to test-leakage, WN18, succeeded by WN18RR~\cite{dettmers2018convolutional}, or CoDEx~\cite{safavi2020codex}, among others. Open-world KGC combines KGs with textual descriptions of its entities. Approaches include the FB20K benchmark as part of DKRL~\cite{xie2016representation}, DBPedia50k and DBPedia500k as part of ConMask~\cite{shi2017open}, and FB15k237-OWE as part of OWE~\cite{shah2019open}. More recent work introduces Wikidata5M~\cite{wang2021kepler} (a large Wikidata subset) and the FB15k237/WN18RR adaptions of~\cite{daza2021inductive}.
Since data in industrial use cases is not publicly available, we also sample our benchmark data from open knowledge graphs (CoDEx) and text sources (Wikipedia). However, all the above benchmarks tackle the open-world scenario with a single concise description of graph entities. We are, however, interested in linking entities associated with many, noisy text contexts. To our knowledge, the only benchmark to attempt this is IRT~\cite{hamann2021open}. Here, a 
a set of $30$ text contexts with incidental mentions is associated with each entity in the knowledge graph. We extend this benchmark by offering more text, multiple graph sizes, a greater variety of mentions, and evaluation protocols for both ranking and linking.

\section{Problem Description}
\glsreset{kg}

We define a \gls{kg} as a directed graph \( \mathcal{G} = (\mathcal{V}, \mathcal{R}, \mathcal{T}, \mathcal{M}, \mathcal{C}) \) where \(\mathcal{V}\) is the set of vertices \(v\) (e.g. \( v=\ktzentity{Fargo} \)) and \( \mathcal{R} \) a set of relation types (e.g. \( r=\ktzrelation{genre} \)). Triples \( (h,r,t) \in \mathcal{T} \subseteq \mathcal{V} \times \mathcal{R} \times \mathcal{V} \) connect the vertices, e.g. (\ktzentity{Fargo}, \ktzrelation{genre}, \ktzentity{Thriller Film}). Additionally, the set \( \mathcal{M} \) contains textual mentions, each  associated with a vertex via a function \( M: \mathcal{V} \mapsto \mathcal{P}(\mathcal{M}) \). For example, the entity v=\ktzentity{Thriller Film} has the mentions M(v) = \{ \ktzmention{crime thriller}, \dots, \ktzmention{gangster film} \}. Note that mentions can be ambiguous (e.g. \ktzmention{the film} as mention of \ktzentity{Fargo}). Furthermore, each of the mentions \( m \) is assumed to occur in textual context sentences \( c \) from a corpus \( \mathcal{C} \). In general, these contexts do not express triples but solely contain incidental mentions of entities.
They are accessible through a function \( C: \mathcal{V} \times \mathcal{M} \mapsto \mathcal{P}(\mathcal{C}) \). For example, with \( v=\ktzentity{Crime Film} \) and \( m=\ktzmention{gangster film} \): \( C(v, m) = \{ \) \ktzsequence{Corman called it the most accurate, authentic \ktzmentionI{gangster film}} \(, \dots \} \). Note that -- via M and C -- we can access an entity's mentions and contexts, and vice versa.

\textbf{Open/Closed-World:}
While the text corpus \( \mathcal{C} \) contains known (or  "closed-world") entities, a second corpus \( \mathcal{Q} \) contains  undiscovered open-world mentions \( \mathcal{M}^o \). Our goal is to discover these and link them to the graph. We define a closed-world graph \( \mathcal{G}^c \) (containing subsets of the respective sets of \( \mathcal{G} \) and the text corpus \( \mathcal{C} \)) and an open-world graph \( \mathcal{G}^o = (\mathcal{V}^o, \mathcal{R}, \mathcal{T}^o, \mathcal{M}^o, \mathcal{Q}) \) with \( \mathcal{V}^o \subseteq \mathcal{V} \), \( \mathcal{T}^o \subseteq \mathcal{T} \), \( \mathcal{M}^o = \mathcal{M} \setminus \mathcal{M}^c \): Namely a graph with a set of undiscovered mentions not known in the closed-world graph \( \mathcal{G}^c \). This gives rise to the two tasks we address:

\textbf{Ranking Task:}
The first task is to retrieve text contexts from \( \mathcal{Q} \) which contain mentions of interest, i.e. unknown mentions of (possibly unseen) entities which should complete a given entity-relation pair. For example, for relation \( r=\ktzrelation{genre} \) and tail \(t=\ktzentity{crime film} \), we search for text contexts where \textit{any} crime films are mentioned.

\textbf{Linking Task:}
Now, given an open-world mention was just discovered and marked, all text contexts of \( \mathcal{Q} \) are bundled by mention. For each such bundle of text, links \( r \in \mathcal{R} \) to the graph vertices \( V^c \) must be found. For example: \ktzsequence{What genre is associated with all text contexts that contain the mention \ktzmentionI{fargo}?}.

\vspace{-1em}
\section{Models}
\label{sec:models}

We tackle the above tasks with three models: (1) \textbf{JOINT}, where a text encoder and a \gls{kgc} scorer are trained jointly, (2) \textbf{OWE}, where an encoder learns to project text representations into a pre-trained graph embedding space, and (3) \textbf{BOW}, a model that employs bag-of-words representations for text similarity. All three models are evaluated on both the ranking and linking tasks. Note that, although our notation is (for brevity) generally focused on finding tails, we do also always include head-prediction for given relation-tail pairs.

\vspace{-.5em}
\subsection{Neural Models}
\begin{figure}[t]
    \centering
    \includegraphics[width=1\textwidth]{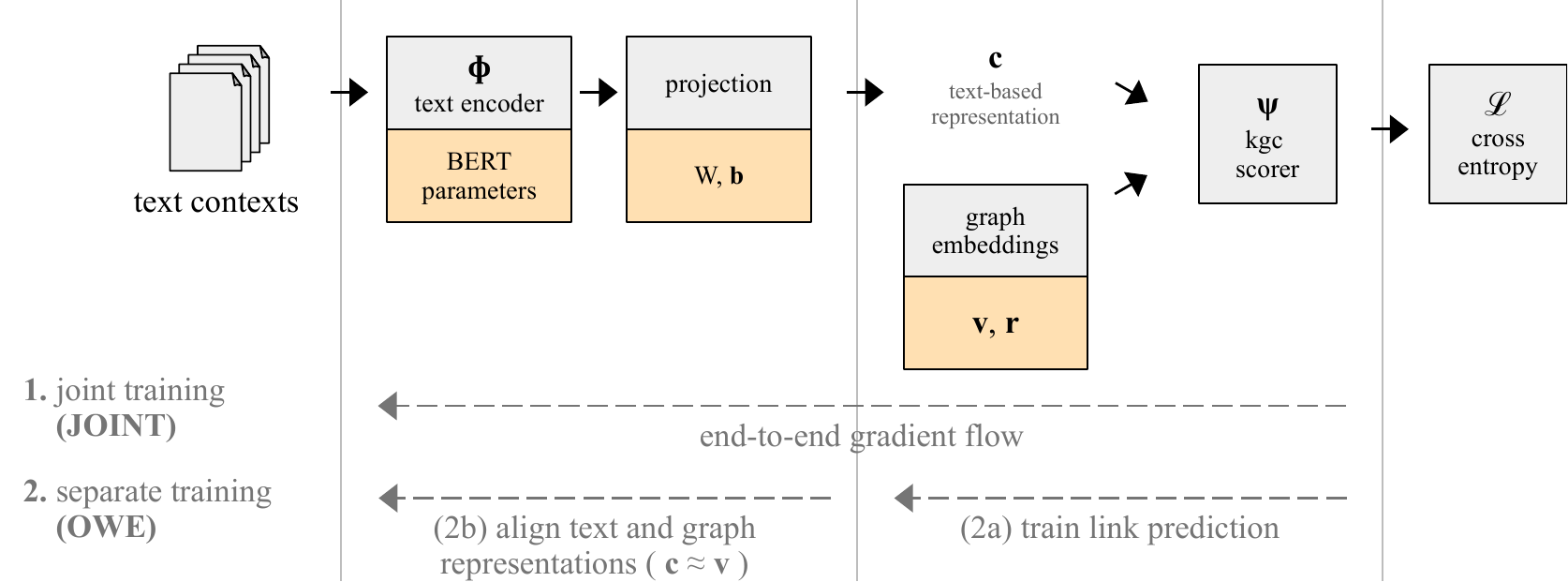}
    \caption{Data flow in our model (top) and the two training strategies (bottom): Strategy 1 (JOINT) applies an end-to-end training of both link predictor and text encoder. Strategy 2 (OWE) first trains the link predictor (2a), and then aligns the text-based entity representations $\mathbf{c}$ with the graph-based ones $\mathbf{v}$ (2b). Learnable parameters are highlighted in orange. }
    \label{fig:models}
\end{figure}

\vspace{-.5em}
Both neural models combine a text encoder \( \phi \) and a \gls{kgc} module \( \psi \) (see \Cref{fig:models}). 
As a text encoder, we use a pre-trained BERT~\cite{devlin2018bert}, a popular state of the art transformer~\cite{vaswani2017attention} encoder. Given a text context \( c \), we run it through the encoder
and select the [CLS]-token embedding as the context representation $\phi(c)_{\text{CLS}} \in \mathbb{R}^{d'}$ (following the observations of~\cite{hamann2021open}). This representation is mapped using a learned affine projection (\( W, \mathbf{b} \)) into a complex-valued space, obtaining a representation \( \mathbf{c} \). The representation is then used to estimate a tripel's plausibility. 
We use the  ComplEx scorer~\cite{trouillon2016complex}, which, given complex-valued  context, relation and tail embeddings \( \mathbf{c}, \mathbf{r}, \mathbf{v} \in \mathbb{C}^d \), calculates the plausibility score as \( \psi(c, r, v) = \textbf{Re}(\mathbf{c} \odot \mathbf{r} \odot \mathbf{\overline{v}}) \). 
Note that---since the projection maps from real-valued to complex space---we choose the matrix and bias vector dimensions accordingly ($W \in \mathbb{R}^{d' \times 2d}, \mathbf{b} \in \mathbb{R}^{2d}$).
Overall, our neural models can be written as:
\begin{equation}
    \label{eq:forward}
s(c, r, v) = \psi( \; \underbrace{W \cdot \phi(c)_{\text{CLS}} + \mathbf{b}}_{\mathbf{c}}, \mathbf{r}, \mathbf{v} \; )
\end{equation}
\label{eq:jnt}

\vspace{-1.5em}
\noindent {\bf Multi-Context Extension}
As discussed above, instead of a single context $c$, a set  $\Sigma$ of {\it multiple} contexts may be available describing the same entity. We would like our entity representations to take the whole set $\Sigma$ into account. 
To do so, we use a simple early fusion that averages the text representations before projecting them:
\begin{align}
\label{eq:multi}
    \mathbf{c} &= W \cdot \Bigg( \frac{1}{|\Sigma|} \sum_{c \in \Sigma} \mathlarger{\phi}(c)_{\text{CLS}} \Bigg) + \mathbf{b} & 
\end{align}
We then use the resulting representation $\mathbf{c}$ as a drop-in replacement in Equation \eqref{eq:forward}. Concerning training, we use two variants of this setup: In the first variant (referred to as \textbf{single}), training happens on individual contexts as in Equation \eqref{eq:forward}, but we apply Equation \eqref{eq:multi} in inference. In the second variant (\textbf{multi}), we apply Equation \eqref{eq:multi} both in training and inference.

\textbf{Training}

As illustrated in Figure \ref{fig:models}, we study two different training strategies to fit the model parameters (text encoder $\phi$, projection $W, \mathbf{b}$, and  knowledge graph embeddings $\mathbf{r}, \mathbf{v}$). 
 Our first approach uses a joint (end-to-end) training (\textbf{JOINT}): We iteratively sample a random entity $v$ from the knowledge graph. For $v$, we draw a random triple $(v,r,v')$ and context $c$.
The model's scores \( s(c, r, v') \) are mapped to a probability distribution using a softmax, and we apply a cross-entropy loss:

\begin{equation}
    \mathcal{L}_{\text{JOINT}} = - \sum_{v' \in \mathcal{V}^c} \Bigg( \log \frac{\exp(s(c, r, v'))}{\sum_{u' \in \mathcal{V}^c} \exp(s(c, r, u')) } \Bigg) \cdot \mathbf{1}_{(v,r,v') \in  \mathcal{T}}
\end{equation}
\label{eq:jnt-loss}

Note that -- though the above notation samples triples $(v,r,v')$ to predict tails $v'$ -- we also sample triples $(v',r,v)$ and predict the heads $v'$ accordingly.

Our second approach follows Shah et al's {\bf OWE} model~\cite{shah2019open} and applies a {\bf separate training}. Here, training happens in two steps: First, the link prediction model \( \psi \) is trained on the closed-world graph, obtaining a {\it graph-based embedding space}.  Second, the text encoder \( \phi \) and projection $(W,\mathbf{b})$ are trained to align entities' text-based representations with the (now fixed) graph-based ones. 
To do so, we reduce the geometric distance between context representations $\mathbf{c} = W \cdot \phi(c)_{CLS} + \mathbf{b}$ and their associated graph-based representations, using a mean squared error loss: 

\begin{equation}
    \mathcal{L}_{\text{OWE}} = \frac{1}{2d} \cdot \sum_{0 < i \leq d} 
    \big(\textbf{Re}(\mathbf{c}_i) - \textbf{Re}(\mathbf{v}_i)\big)^2
    + 
    \big(\textbf{Im}(\mathbf{c}_i) - \textbf{Im}(\mathbf{v}_i)\big)^2
\end{equation}
\label{eq:owe-loss}

\textbf{Inference:} This section outlines how the neural models are used in our two tasks. 
First, in \textbf{ranking}, a closed-world entity \(v\) and a relation of interest \( r \) are given, and the task is to retrieve a ranked list of text contexts from the query corpus \( \mathcal{Q} \) that potentially contain mentions of relevant open-world entities to be identified by the expert. To do so, we compute the single-context scores $s(c,r,v)$ for all contexts $c \in Q$, $ r \in \mathcal{R}$ and $ v \in \mathcal{V}^c $. The scores are normalized per relation using the softmax function as to better compare the independently scored contexts $c$ (the raw ComplEx scores are unbounded). The contexts in the query corpus are then ranked by this normalized score.

Second, in \textbf{linking}, the expert is interested in a certain mention of interest  \( m \in \mathcal{M}^o \) representing an open-world entity. His/her goal is to link said entity to the graph. To do so, he/she selects a relation \( r \in \mathcal{R} \) and collects a set of contexts $\Sigma$ containing $m$ to describe the open-world entity.
Using the contexts, all entity candidates \( v \in \mathcal{V}^c \) are ranked by their scores \( s(\Sigma, r, v) \). Note that in principle we can repeat this procedure for each relation and also for predicting heads instead of tails.

\subsection{BOW: Bag-of-Words Retrieval}

To evaluate the tasks at hand with a common industrial approach, we compare the above neural models with a bag-of-words approach. Given one or several mentions, this approach concatenates text contexts containing the mentions into {\it documents} and then conducts a similarity matching between documents using the well-known BM25\cite{robertson1994some} keyword matching implemented by the Elasticsearch search engine\footnote{https://www.elastic.co}.

\textbf{Ranking:} Given an entity relation pair such as \( v=\ktzentity{Thriller Film} \), \( r=\ktzrelation{genre} \) (``Find text contexts which mention thriller films''), the baseline's strategy is to find open-world entities that are similar to known closed-world entities $v'$ for which $(v',r,v)$ holds (i.e., known thriller films). We sample a set of random \textit{representatives} \( v' \) and some of their associated text contexts. These contexts are concatenated into a document which is then used as a query against an index containing all open-world contexts \( \mathcal{Q} \).

\textbf{Linking:} Given a mention \( m \in \mathcal{M}^o \) and a relation \( r \), we sample contexts containing $m$ into a document \( D(m) \), which is used to query against an index containing documents \( D(v) \) for all closed-world vertices \( v \in \mathcal{V}^c \). The top-n predicted results allow us to compile a set of reference vertices which we use as blueprints for prediction. For example: When searching for \( r = \ktzrelation{profession} \) of text contexts containing \ktzmention{s. a. corey} (\ktzsequence{Predict the profession of the text contexts containing \ktzmentionI{s. a. corey}}), the result list may contain other texts describing authors. The final prediction of vertices is compiled from the retrieved vertices targets of relation \( r \) and scored by the inverse of the position of the document in the result list.

\section{Benchmark Construction}
\label{sec:benchmark}

Our benchmark picks up the work of~\cite{hamann2021open} and extends it: (1) We study the behaviour of models with varying knowledge graph size, and particularly for small graphs, (2) we test models for both the ranking and linking task. We offer four variants of the benchmark: \textbf{tiny}, \textbf{small}, \textbf{medium}, and \textbf{large}. The variants offer different ratios of structural information (i.e. graph triples) and associated text. The datasets are based on CoDEx-M~\cite{safavi2020codex} which is sampled from Wikidata~\cite{vrandevcic2014wikidata}. The following steps are executed:

\textbf{1. Identifying Concept Vertices:} We assume that in a real-world scenario \textit{world-knowledge} is more likely to be already modelled in the given \gls{kg}. Information to be discovered is usually much more volatile. Consider for example a graph for a machine manufacturer. We know that things can catch fire (the world knowledge) but we aim to find the specific parts that actually do. A heuristic to identify such world knowledge is based on the assumption that such entities are characterised by relations which exhibit a strong disproportion between their heads and tails. The ratio between a relation \( r\)'s domain size (its number of heads) \( \text{dom}(r) := |\{h \:|\: \exists\: t: \: (h, r, t) \in \mathcal{T}\}| \) and the range size (its number of tails) \( \text{rg}(r) := |\{t \:|\: \exists\: h: \: (h, r, t) \in \mathcal{T} \}| \) is \( \text{min}\big(\text{dom}(r), \text{rg}(r)\big) / \text{max}\big(\text{dom}(r), \text{rg}(r)\big) \). Per size variant of the dataset, we select a subset of the relations offered by CoDEx, order by ratio and select a subset of those to determine \textit{concept entities} that will remain in the closed-world split. The appendix in \Cref{sec:appendix} enumerates all relations and corresponding selections in greater detail.

\textbf{2. Sampling Mentions and Text Contexts:} 
As our graph is based on Wikidata, we can exploit its links to Wikipedia. For each vertex \( v \), we identify the associated Wikipedia page \( p_v \) and all pages which link back to it \( \mathcal{N}(p_v) \). Using the link text of the hyperlinks, we build the set of mentions which are associated with the vertex \( M(v) \). Lastly, to build the contexts \( C(v, m) \), all occurrences of the mentions in \( \mathcal{N}(p_v) \) are identified and the surrounding sentence kept as context. This is a heuristic to obtain incidental text contexts: The entity is mentioned but usually not the \textit{subject} of the sentence. As example for \( v = \ktzentity{Scotland} \): \ktzsequence{Aberdeen is a city in northeast \underline{Scotland}}.

\textbf{3. Open/Closed-World Split:}
The open/closed-world split, in this case, is not done on vertex but mention level. We separate the mention set \( \mathcal{M} \) into a partition \( \mathcal{M}^c \) for closed-world- and \( \mathcal{ M}^o \) for open-world mentions respectively. First, all mentions associated with concept entities are set aside for the closed-world part. Then, for each remaining vertex \( v \), its associated mentions are distributed randomly between open- and closed-world. We set an additional pruning parameter which limits the maximum amount of mentions allowed for a closed-world vertex. With the information about closed-world mentions, we identify all triples whose vertices are associated with them and set them aside as the closed-world triple set \( \mathcal{T}^c \subseteq \mathcal{T} \). The so-called \textit{test triples} \( (m \in \mathcal{M}^o, r \in \mathcal{R}, v \in \mathcal{V}^c) \) (explained below the table) are derived directly from the triple set \( \mathcal{T} \) by identifying in which triples the mention's vertex occurs. We set aside a subset of the open-world split for validation. The above steps are executed for the four different dataset variants. These have different parameters set for concept relations, total relations to keep and closed-world mention pruning (see the appendix for more details).

{\footnotesize
\begin{center}
\begin{tabular}{ll|rrrrr}\toprule
    &\quad\quad&\quad\quad& \textbf{Tiny} & \textbf{Small} & \textbf{Medium} & \textbf{Large} \\\hline
    Relations \(|\mathcal{R}| \) &&& 5 & 12 & 45 & 45 \\
    Entities \(|\mathcal{E}| \) &&& 1,174 & 2,887 & 3,592 & 9,952 \\
    \hline
    Training Triples \( |\mathcal{T}^c| \) &&& 2,928 & 7,527 & 26,335 &  102,289 \\
    Training Text \( |\mathcal{C}| \) &&& 9,100,422 & \quad15,117,184 & \quad17,398,943 & \quad18,654,485 \\
    \hline
    Test Text \( |\mathcal{Q}| \) &&& 6,058,557 & 2,390,017 & 1,905,151 & 864,598 \\
    Ranking Queries &&& 695 & 1,225 & 4,972 & 7,336 \\
    Linking Queries &&& 47,857 & 52,654 & 97,921 & 48,801 \\
    Test Triples &&& 102,866 & 102,616 & 174,312 & 90,780 \\
    \bottomrule
\end{tabular}
\end{center}
}

The last four rows of the table describe the scale of the challenges. A \textbf{ranking query} is a tuple \( (v \in \mathcal{V}^c, r \in \mathcal{R}) \) (e.g. (\ktzentity{writer}, \ktzrelation{profession}) -- \textit{``Find texts with writers in them''}) and an associated ground truth set of open-world mentions \( \{ m_1, \dots \} \in M^o \) to be discovered (e.g. \{ \ktzmention{Tolkiens}, \ktzmention{the author}, \ktzmention{Corey}, \dots \}). A \textbf{linking query} is a tuple \( (m \in M^o, r \in \mathcal{R}) \) (e.g. (\ktzmention{Tolkiens}, \ktzrelation{profession}) -- ``What professions does the entity have, given texts in which \ktzmention{Tolkiens} occurs?'') and a set of associated closed-world vertices \( v \in \mathcal{V}^c \) (e.g. \{ \ktzentity{author}, \ktzentity{linguist}, \dots \}). One can see that both ranking and linking are different views of the same data: unique \( (m \in M^o, r \in \mathcal{R}, v \in \mathcal{V}^c) \) triples. These triples are presented as \textit{test triples} in the table above and allow to gauge the ratio of the queries and their associated ground truth. For example, in the most extreme case, ranking on the tiny variant, has \(102866 / 695 \approx 148 \) mentions to discover per query on average.

\section{Experiments}
\label{sec:experiments}

We run a series of experiments to build a solid understanding of the challenge's difficulty. We employ all three formerly described models (JOINT, OWE, BOW) for both tasks (ranking, linking) on all four splits (tiny, small, medium, and large). We measure hits@k and the mean reciprocal rank (MRR) for each test triple (i.e. micro-averaged). Following common practice, we apply target filtering~\cite{shah2019open} to the scored predictions (mentions for ranking and vertices for ranking). Target filtering is a method where, for a given ranking, only a single true positive of interest is inspected, while all other true positives are removed from the list. This helps to not artificially worsen the result for rankings with many true positives. We offer the datasets, dataset creation code, and evaluation protocol online\footnote{\url{https://github.com/lavis-nlp/irt2}}. The model implementations and all presented trained models are also supplied separately\footnote{\url{https://github.com/lavis-nlp/irt2m}}. We evaluate our models on a subset of the test text contexts \( \mathcal{Q} \) (400k sentences drawn randomly for ranking; 100 sentences per mention for linking).

\subsection{Hyperparameters}
\label{sec:hyperparameters}

We determine the final hyperparameters using a combination of random-, and grid searches over the parameter space. A single closed-world ComplEx model is trained per dataset split and commonly used to train the \textbf{OWE} model using Adagrad~\cite{duchi2011adaptive}. The hyperparameters studied comprise learning rate, L2 regularization weight, and embedding space dimensions. We employ PyKEEN\cite{ali2020pykeen} for training and build our own open-world extension. We implement and train the \textbf{JOINT} and \textbf{OWE} models using PyTorch~\cite{paszke2019pytorch} and PyTorch Lightning\cite{falcon2019lightning}. Models are optimized using Adam\cite{kingma2014adam}. For each model architecture and dataset split, we run a random parameter sweep to fix a subset of parameters and subject the remaining to a grid search. We study the effect of regularization of the entity and relation embeddings, weight-decay and learning rate of the optimizer, how many layers of the encoder are frozen during training, how many text contexts are sampled per vertex per training, and whether the mention is masked during training. Additionally, we study how many contexts to provide per optimizer step for multi-context models. Generally, all models share a similar set of values for learning-rate, weight-decay and regularization weight. The other parameters are dependent on the split. We found to freeze parts of the encoder works well against overfitting. Contrary to the observations of~\cite{hamann2021open}, masking did not significantly increase model performance. The final hyperparameters are provided in the appendix (\Cref{sec:appendix}) and with the models online.

\subsection{Linking}

First, we evaluate the linking task as the designated challenge for the presented models. \Cref{tab:linking} details our findings. Generally, a few trends can be observed: (1) The neural models significantly outperform the baseline approach on the three smaller variants but yield inferior results on the large dataset. This supports our claim that a benchmark should offer insights into varying degrees of structural data scarcity. Even with an abundance of data (usually favouring neural approaches), the BOW baseline can beat the neural model's performance on large but falls short for the smaller variants. (2) Overall, the OWE approach outperforms JOINT albeit not by a great margin for tiny, small, and large. The biggest performance gap can be observed on medium, where the OWE model profits the most from the decoupled training. We argue that this hints towards the need for better sampling/overfitting strategies during training for JOINT as it seems to struggle with the lower variety of mentions seen during training. (3) Although the multi-context models generally perform better than the single-context counterparts, the difference in performance is not as pronounced as reported by~\cite{hamann2021open}. Contrary to their findings we also did not observe much performance improvement through masking the mention. These observations indicate that both models struggle to incorporate much of the textual clues and instead rely heavily on the structural information present in the vertex- and relation embeddings. Overall, the metrics show that the models can link an entity described by text contexts to a given graph with relatively high precision.

\begin{table}[!htp]\centering
\begin{tabular}{ll|rrrr|rrrr}\toprule
& &\multicolumn{4}{c}{\textbf{HITS@10}} &\multicolumn{4}{c}{\textbf{MRR}} \\\cmidrule{3-10}
& &\textbf{Tiny} &\textbf{Small} &\textbf{Med.} &\textbf{Large} &\textbf{Tiny} &\textbf{Small} &\textbf{Med.} &\textbf{Large} \\\midrule
BOW & &53.82 &55.18 &46.43 &\textbf{71.38} &33.63 &34.62 &29.81 &\textbf{50.61} \\
JOINT &single &72.06 &70.20 &47.14 &65.75 &50.61 &45.95 &33.72 &48.29 \\
JOINT &multi &73.56 &74.27 &53.77 &65.12 &51.28 &\textbf{52.39} &37.50 &45.26 \\
OWE &single &74.09 &\textbf{74.33} &61.98 &64.27 &50.25 &50.57 &40.60 &42.69 \\
OWE &multi &\textbf{75.39} &71.49 &\textbf{64.41} &66.36 &\textbf{53.06} &47.17 &\textbf{43.25} &45.51 \\
\bottomrule
\end{tabular}

\caption{Model performance on the \textbf{linking} task. The bag-of-words model outperforms the neural approaches on the large dataset. The neural models both outperform the baseline on the smaller variants by a large margin. Among the neural models, OWE generally yields better performance on all splits.}
\label{tab:linking}
\end{table}

\subsection{Ranking}

Secondly, for ranking, we study whether the models which were originally trained to do link-prediction can be employed for discovery, too. We report the hits@100 performance in \Cref{tab:ranking}. (1) The results show that the scoring mechanism of the neural models outperforms the BM25 search results. At first glance, this is a surprising insight, but can be explained by the nature of text samples. As the mention of interest is usually not the subject of the text context, the query does not directly describe its properties. This leads the ranker astray. For example: searching for other comedians (i.e. ?, \ktzrelation{profession}, \ktzentity{comedian}) the model uses text samples sampled from the Wikipedia page about New York among others. This leads to other text samples to be retrieved which also talk about things in the context of New York but seldom other comedians. This \textit{contextualization} of queries (by queried relation) is better considered with the neural models. (2) Overall ranking quality is not high enough for practical application.

\begin{table}[!htp]\centering
\begin{tabular}{ll|rrrr}\toprule
& &\multicolumn{4}{c}{\textbf{HITS@100}} \\\cmidrule{3-6}
& &\textbf{Tiny} &\textbf{Small} &\textbf{Med.} &\textbf{Large} \\\midrule
BOW & &2.86 &4.29 &6.42 &14.83 \\
JOINT &single &7.91 &6.78 &6.37 &19.47 \\
JOINT &multi &\textbf{13.28} &\textbf{16.17} &\textbf{14.38} &30.68 \\
OWE &single &6.30 &8.19 &6.88 &10.81 \\
OWE &multi &9.98 &13.00 &6.36 &\textbf{31.40} \\
\bottomrule
\end{tabular}

\caption{Results for the \textbf{ranking} task. The neural models all outperform the baseline and, contrary to the linking task, the multi-context JOINT model is generally the best performing approach.}
\label{tab:ranking}

\end{table}

We suspect two main factors which hinder effective ranking: First, the scores are calculated independently from each other. This makes an intra-sample comparison of scores difficult. Secondly, the ranking task requires scores first and foremost produced by head prediction. These scores usually are of lower quality because the models are much better at predicting a few biased tails (e.g. professions) than hundreds or thousands of volatile heads (e.g. authors). This is worsened by the formerly described observation that text contexts have a comparatively small influence on the score. To obtain better performance, new approaches must be devised which incorporate the graph context on the one hand and neural semantic retrieval~\cite{mitra2017neural,lin2021pretrained} on the other.

\section{Conclusion}

We present IRT2, a benchmark to study a knowledge acquisition pipeline to extend a knowledge graph (KG) from noisy text. We offer knowledge graphs of different sizes and text associated with its vertices. The benchmark comprises two tasks, \textit{ranking} and \textit{linking}, where first, entities need to be discovered from unlabelled text, and in the second stage, linked to the KG. Alongside, we study three approaches for the tasks at hand: two neural open-world triple scorers and a bag-of-words model. We found the models to perform well for the linking task but less so for ranking. For linking, the BOW baseline outperforms the neural models on the large variant. For the remaining, smaller variants, the OWE approach outperforms all other approaches even if graph data is scarce. For ranking, the JOINT model produces the best results. Although the models show a promising direction to be used for ranking, they do not perform at the level for practical application yet. We recommend studying how semantic ranking techniques can profit from the given links between graph- and text data. We invite the community to use the benchmark to (1) devise ways to better incorporate the noisy but abundant text data to make better predictions, (2) find ways to contextualize semantic retrievers with the specific graph information, and (3) derive insights for application in an industrial context with similar constraints and conditions. The dataset and evaluation scripts can be found here: \url{https://github.com/lavis-nlp/irt2}. 

\textbf{Acknowledgement:} This work was supported by the BMBF program FH-Kooperativ, project SCENT (13FH003KX0).

%

\bibliography{literature}

\newpage

\section{Appendix}
\label{sec:appendix}

This appendix details configuration options used for benchmark construction and model training in detail to allow for the reproduction of the reported results.

\subsection{Benchmark}

\begin{tabular}{lrrrrr}\toprule
\textbf{Split} &\textbf{Tiny} &\textbf{Small} &\textbf{Medium} &\textbf{Large} \\\midrule
Concept Relations &4 &8 &27 &27 \\
Total Relations &5 &12 &45 &45 \\
Closed World Threshold &400 &800 &800 &- \\
Target Mention Split &70\% &70\% &70\% &70\% \\
Target Validation Split &10\% &20\% &20\% &80\% \\
Mention Threshold &5 &5 &5 &5 \\\hline
Concept Vertices &471 &962 &2,200 &2,200 \\
Training Vertices &1,174 &2,887 &3,592 &9,952 \\
Training Mentions &4,945 &9,231 &14,417 &22,866 \\
Training Triples &2,928 &7,527 &26,335 &102,289 \\
Training Text Contexts &9,100,422 &15,117,184 &17,398,943 &18,654,485 \\\hline
Validation Vertices &1,741 &3,375 &3,279 &2,644 \\
Validation Mentions &1,894 &3,870 &3,649 &2,940 \\
Validation Vertex Triples &13,639 &24,764 &43,532 &35,209 \\
Validation Task Triples &11,588 &25,722 &43,716 &38,582 \\
Validation Text Contexts &850,351 &597,050 &505,979 &383,351 \\\hline
Test Vertices &10,909 &10,527 &10,221 &5,607 \\
Test Mentions &17,055 &15,481 &14,600 &6,860 \\
Test Vertex Triples &70,094 &71,360 &124,050 &71,951 \\
Test Task Triples &102,866 &102,616 &174,312 &90,780 \\
Test Text Contexts &6,058,577 &2,390,017 &11,705,151 &864,598 \\
\bottomrule
\end{tabular}
\vspace{1em}

The final benchmark proportions rely heavily on the selected relations and pruning options for construction. The following table enumerates both the configuration which was used for sampling and the resulting key figures for the training, validation and test splits. Relations are picked manually per split, are ordered by their ratio and a subset is selected as concept relations. The \textit{Closed World Threshold} determines how many mentions per relation are retained at most in the closed world split. The \textit{Mention Threshold} says to keep only mentions with at least that many associated text contexts. \textit{Validation-} and \textit{Test Vertex Triples} are the triple sets (vertex, relation, vertex) from which the final \textit{Task Triples} are derived (mention, relation, vertex). They are pruned by filtering out \textit{true} open-world vertices (the necessary zero-shot entity linking is out of scope for the presented tasks)---which explains the decrease that can be observed for the tiny variant.

{\begin{center}
\includegraphics[width=.7\textwidth]{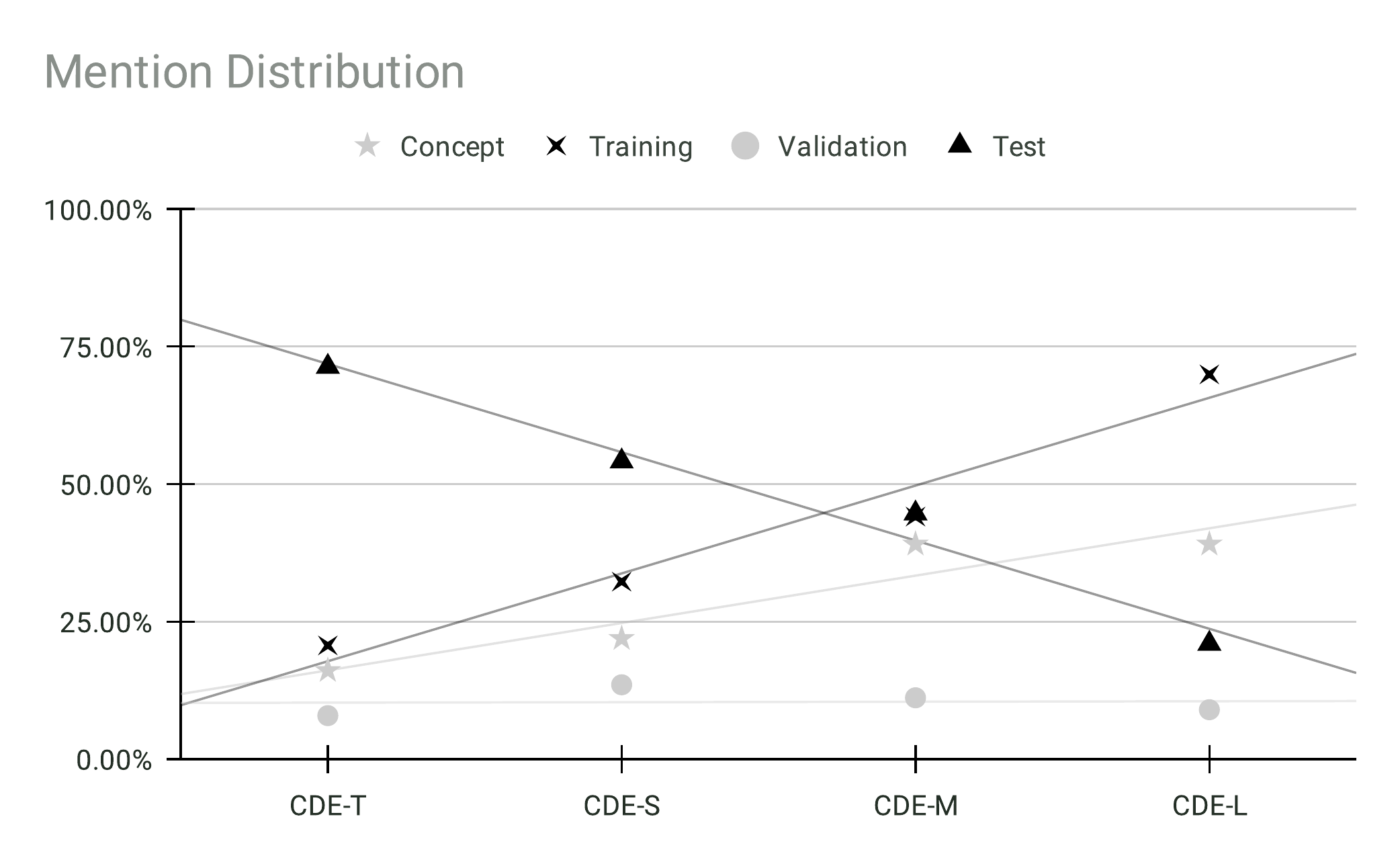}\\
\includegraphics[width=.7\textwidth]{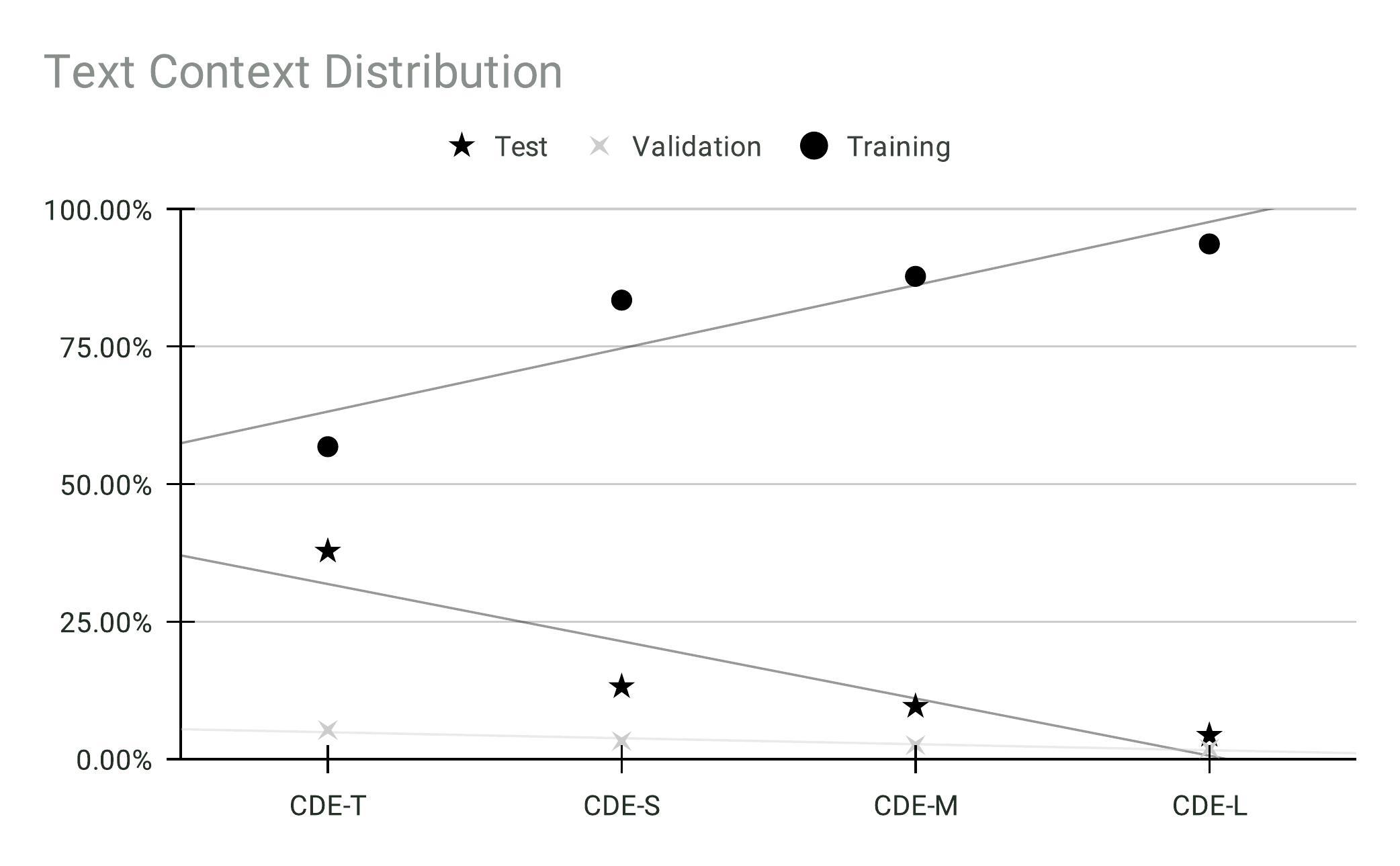}
\end{center}}

The above figures detail the shift in proportions between the dataset variants. The test mention counts decrease proportionally to the increase in training. Concept mention count increases and is equal for M and L. The proportion of concept mentions decreases with increasing dataset size. Correspondingly, validation text context counts stay constant while an increase in training data decreases available test data.

{\begin{center}
\includegraphics[width=.7\textwidth]{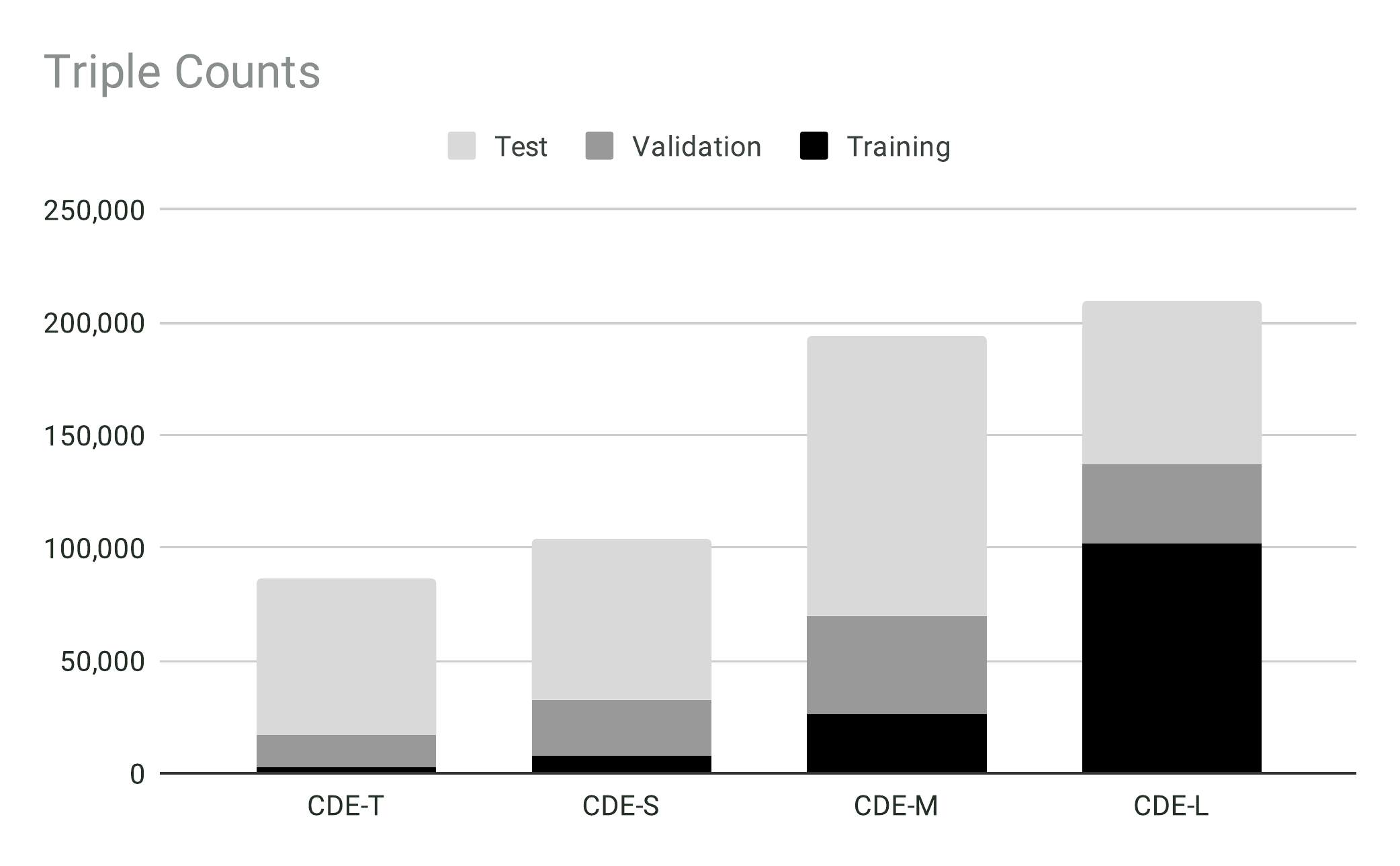}
\includegraphics[width=.7\textwidth]{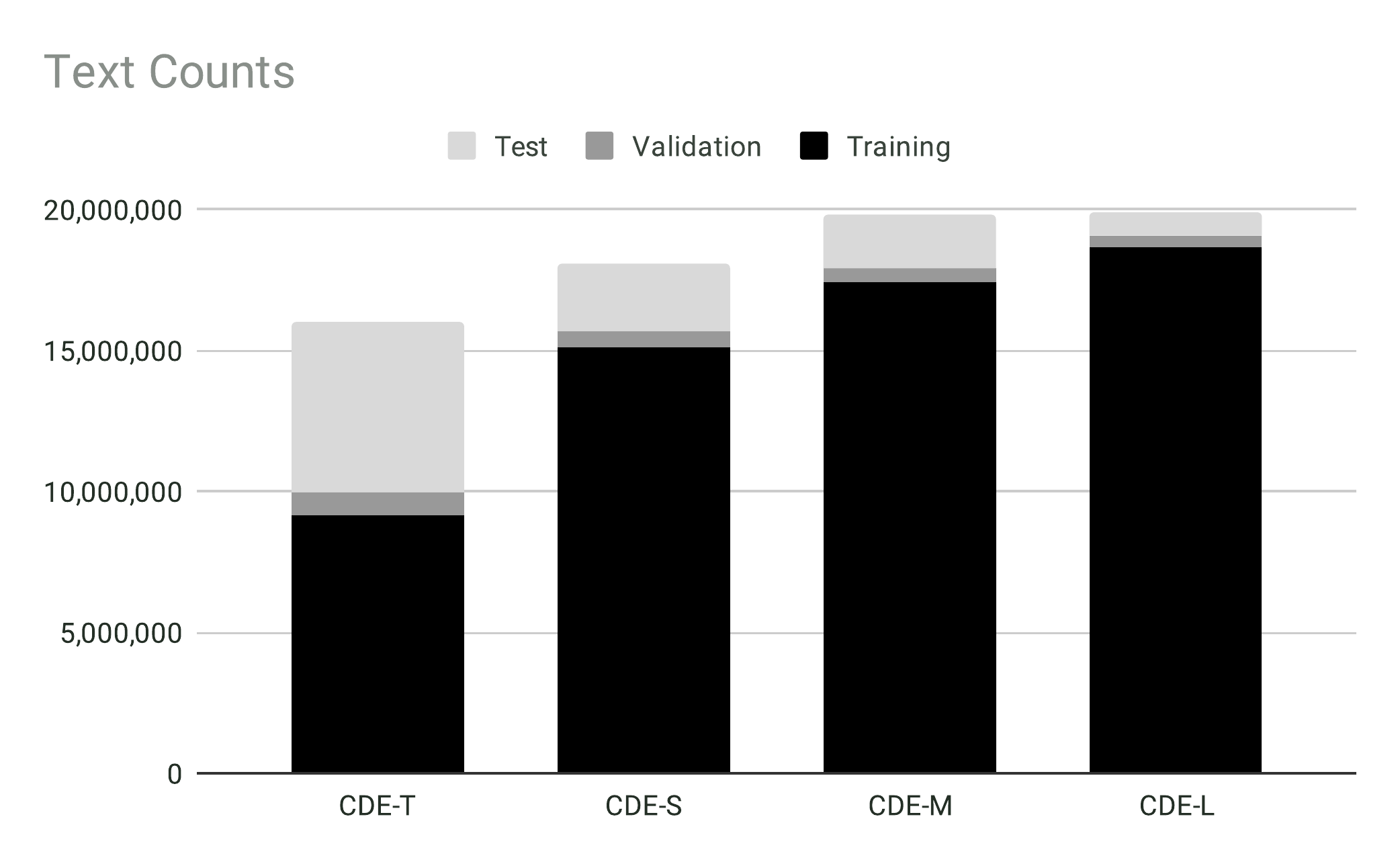}
\end{center}}

The bar plots above report the absolute counts of triples (called \textit{Vertex Triples} in the former table) and associated text. The resulting triple split behaves correspondingly to the mention split with a proportional behaviour between training/test data. The overall triple count more than doubles between T and L. Associated text data increases with respect to the increase in closed-world mentions.

The following table shows all relations present in CoDEx ordered by ratio as described in \ref{sec:benchmark}. On the right-hand side, it is detailed which relations are present in which dataset variant. The respective right side shows whether it is selected to remain in the dataset. The left side marks whether the relation was selected to pick concept entities.

{\footnotesize
\begin{adjustbox}{width=\textwidth}
\begin{tabular}{rl|rrrr|p{.25cm}p{.25cm}|p{.25cm}p{.25cm}|p{.25cm}p{.25cm}|p{.25cm}p{.25cm}}\toprule
\textbf{ID} &\textbf{Relation} &\textbf{Ratio} &\textbf{Heads} &\textbf{Tails} &\textbf{Triples} &\multicolumn{2}{c}{\textbf{T}} &\multicolumn{2}{c}{\textbf{S}} &\multicolumn{2}{c}{\textbf{M}} &\multicolumn{2}{c}{\textbf{L}} \\\midrule
P1412 & languages spoken (...) &6.32e-3 &9,816 &62 &12,584 &X &X &X &X &X &X &X &X \\
P1303 & instrument &1.05e-2 &3,622 &38 &6,076 & & & & &X &X &X &X \\
P140 & religion &1.59e-2 &2,520 &40 &2,651 & & &X &X &X &X &X &X \\
P27 & country of citizenship &1.69e-2 &13,036 &220 &16,828 & &X &X &X &X &X &X &X \\
P30 & continent &1.98e-2 &353 &7 &391 &X &X &X &X &X &X &X &X \\
P509 & cause of death &2.02e-2 &3,071 &62 &3,210 & & & & &X &X &X &X \\
P172 & ethnic group &2.48e-2 &2,013 &50 &2,293 & & & & &X &X &X &X \\
P2348 & time period &2.63e-2 &152 &4 &152 & & & & &X &X &X &X \\
P102 & member of political party &2.76e-2 &2,175 &60 &2,668 & & & & &X &X &X &X \\
P106 & occupation &2.85e-2 &13,145 &375 &71,596 &X &X &X &X &X &X &X &X \\
P495 & country of origin &3.85e-2 &1,299 &50 &2,049 & & &X &X &X &X &X &X \\
P136 & genre &4.32e-2 &5,303 &229 &11,761 & & & & &X &X &X &X \\
P641 & sport &4.43e-2 &384 &17 &418 & & & & &X &X &X &X \\
P19 & place of birth &6.11e-2 &7,185 &439 &7,214 & &X &X &X &X &X &X &X \\
P69 & educated at &6.47e-2 &6,502 &421 &9,752 & & & & &X &X &X &X \\
P463 & member of &6.77e-2 &3,705 &251 &11,490 & & & & &X &X &X &X \\
P264 & record label &6.84e-2 &2,002 &137 &3,456 & & & & &X &X &X &X \\
P20 & place of death &6.96e-2 &5,417 &377 &5,442 & & & & &X &X &X &X \\
P1050 & medical condition &7.59e-2 &395 &30 &408 & & & & &X &X &X &X \\
P101 & field of work &8.13e-2 &1,967 &160 &2,421 & & & & &X &X &X &X \\
P2283 & uses &8.33e-2 &12 &1 &12 & & & & &X &X &X &X \\
P135 & movement &9.20e-2 &413 &38 &446 & & & & &X &X &X &X \\
P119 & place of burial &9.67e-2 &1,944 &188 &1,972 & & &X &X &X &X &X &X \\
P108 & employer &1.27e-1 &3,016 &382 &4,795 & & & &X &X &X &X &X \\
P37 & official language &1.83e-1 &306 &56 &403 & & & & &X &X &X &X \\
P840 & narrative location &1.97e-1 &986 &194 &1,506 & & & & &X &X &X &X \\
P17 & country &2.23e-1 &641 &143 &1,323 & & & & &X &X &X &X \\
P50 & author &2.35e-1 &4 &17 &17 & & & & & & & & \\
P452 & industry &2.50e-1 &16 &4 &17 & & & &X & &X & &X \\
P551 & residence &2.52e-1 &1,426 &359 &1,934 & & & & & &X & &X \\
P749 & parent organization &3.15e-1 &73 &23 &82 & & & & & &X & &X \\
P407 & language of work or name &3.24e-1 &34 &11 &43 & & & & & &X & &X \\
P361 & part of &3.48e-1 &138 &48 &171 & & & & & &X & &X \\
P57 & director &3.71e-1 &542 &201 &561 & & & & & &X & &X \\
P159 & headquarters location &4.20e-1 &157 &66 &169 & & & &X & &X & &X \\
P161 & cast member &4.80e-1 &1,227 &2,557 &9,249 & & & & & &X & &X \\
P1056 & product (...) produced &5.00e-1 &2 &1 &2 & & & & & & & & \\
P740 & location of formation &5.19e-1 &133 &69 &133 & & & & & &X & &X \\
P131 & located in (...) &7.61e-1 &163 &124 &212 & & & & & &X & &X \\
P737 & influenced by &8.71e-1 &514 &590 &1,508 & & & & & &X & &X \\
P138 & named after &8.91e-1 &49 &55 &57 & & & & & &X & &X \\
P112 & founded by &9.48e-1 &55 &58 &67 & & & & & &X & &X \\
P40 & child &9.54e-1 &309 &324 &391 & & & &X & &X & &X \\
P451 & unmarried partner &9.62e-1 &328 &341 &393 & & & & & &X & &X \\
P530 & diplomatic relation &9.68e-1 &214 &221 &6,225 & & & & & &X & &X \\
P3373 & sibling &9.95e-1 &394 &396 &785 & & & & & &X & &X \\
P26 & spouse &1.00e+0 &804 &804 &866 & & & & & &X & &X \\
P3095 & practiced by &1.00e+0 &2 &2 &2 & & & & & & & & \\
P54 & member of sports team &1.00e+0 &2 &2 &2 & & & & & & & & \\
P113 & airline hub &1.00e+0 &1 &1 &1 & & & & & & & & \\
P780 & symptoms &1.00e+0 &1 &1 &1 & & & & & & & & \\
\bottomrule
\end{tabular}
\end{adjustbox}
}

\subsection{Hyperparameters}

All models are trained on Nvidia GTX 2080TI or RTX A6000 graphics cards using PyTorch version 1.11. KGC models use the 1.8 implementation of PyKEEN and the text encoder uses Huggingface's transformer implementation 4.19. The amount of text samples per epoch varies between different model configurations and is a combination of ``max contexts'' (the total amount of text contexts associated with an entity during training) ``max contexts per sample'' (the number of text contexts used by a model per training step). Model training is stopped if the performance did not increase on the target metric (i.e. hits@10) on the validation split by more than 0.001 for a few epochs. In the following, the hyperparameters for the models presented in \ref{sec:experiments} are enumerated.
\vspace{1em}

\begin{table}[!htp]\centering
\scriptsize
\begin{tabular}{l|rrrr|rrrr}\toprule
\textbf{} &\multicolumn{4}{c|}{\textbf{Ranking}} &\multicolumn{4}{c}{\textbf{Linking}} \\\cmidrule{2-9}
&\textbf{Tiny} &\textbf{Small} &\textbf{Medium} &\textbf{Large} &\textbf{Tiny} &\textbf{Small} &\textbf{Medium} &\textbf{Large} \\\midrule
Embedding Dims. &800 &500 &800 &200 &200 &800 &800 &200 \\
Unfrozen Layer &5 &0 &11 &0 &5 &11 &11 &0 \\
Regularizer Weight &7.16e-1 &7.73e-1 &7.12e-2 &5.86e-3 &6.56e-1 &9.02e-3 &7.12e-2 &5.86e-3 \\
Contexts per Sample &1 &1 &1 &1 &1 &1 &1 &1 \\
Maximum Contexts &10 &100 &10 &10 &10 &10 &10 &10 \\
Masked &false &false &false &false &true &true &false &false \\
Batch Size &4 &8 &40 &20 &4 &8 &40 &20 \\
Learning Rate &3.94e-6 &4.93e-6 &4.39e-6 &8.99e-6 &8.47e-6 &6.67e-6 &4.39e-6 &8.99e-6 \\
Weight Decay &3.48e-2 &1.19e-2 &6.64e-4 &1.49e-4 &1.22e-4 &2.16e-4 &6.64e-4 &1.49e-4 \\
Seed &5629275 &4828059 &2773483 &4076657 &8697782 &9387603 &2773483 &4076657 \\
\bottomrule
\end{tabular}
\caption{JOINT (single context) - best model selected after random search.}
\end{table}

\begin{table}[!htp]\centering
\scriptsize
\begin{tabular}{l|rrrr|rrrr}\toprule
\textbf{} &\multicolumn{4}{c|}{\textbf{Ranking}} &\multicolumn{4}{c}{\textbf{Linking}} \\\cmidrule{2-9}
&\textbf{Tiny} &\textbf{Small} &\textbf{Medium} &\textbf{Large} &\textbf{Tiny} &\textbf{Small} &\textbf{Medium} &\textbf{Large} \\\midrule
Embedding Dims. &800 &800 &800 &800 &800 &800 &800 &800 \\
Unfrozen Layer &1 &1 &1 &1 &1 &1 &1 &1 \\
Regularizer Weight &0.01 &0.01 &0.01 &0.01 &0.01 &0.01 &0.01 &0.01 \\
Contexts per Sample &40 &40 &40 &10 &40 &40 &40 &10 \\
Maximum Contexts &100 &100 &1000 &1000 &100 &100 &1000 &1000 \\
Masked &false &false &false &false &false &false &false &false \\
Batch Size &1 &1 &1 &4 &1 &1 &1 &4 \\
Subbatch Size &5 &5 &10 &10 &5 &5 &10 &10 \\
Learning Rate &5.00e-6 &5.00e-6 &5.00e-6 &5.00e-6 &5.00e-6 &5.00e-6 &5.00e-6 &5.00e-6 \\
Weight Decay &0.0001 &0.0001 &0.0001 &0.0001 &0.0001 &0.0001 &0.0001 &0.0001 \\
Seed &4733486 &5969007 &7460979 &6929361 &4733486 &5969007 &7460979 &6929361 \\
\bottomrule
\end{tabular}
\caption{JOINT (multi-context) - best model selected after grid search.}
\end{table}

\begin{table}[!htp]\centering
\scriptsize
\begin{tabular}{l|rrrr|rrrr}\toprule
\textbf{} &\multicolumn{4}{c|}{\textbf{Ranking}} &\multicolumn{4}{c}{\textbf{Linking}} \\\cmidrule{2-9}
&\textbf{Tiny} &\textbf{Small} &\textbf{Medium} &\textbf{Large} &\textbf{Tiny} &\textbf{Small} &\textbf{Medium} &\textbf{Large} \\\midrule
Contexts per Sample &1 &1 &1 &1 &1 &1 &1 &1 \\
Maximum Contexts &1 &1 &1 &1 &1 &1 &1 &1 \\
Masked &false &false &false &false &false &false &false &false \\
Batch Size &10 &10 &10 &10 &10 &10 &10 &10 \\
Subbatch Size &10 &10 &10 &10 &10 &10 &10 &10 \\
Learning Rate &5.00e-5 &5.00e-5 &5.00e-5 &5.00e-5 &5.00e-5 &5.00e-5 &5.00e-5 &5.00e-5 \\
Seed &1705119 &2298300 &3899079 &8847385 &1705119 &2298300 &3899079 &8847385 \\
\bottomrule
\end{tabular}
\caption{OWE (single context) - best model selected after grid search.}
\end{table}

\begin{table}[!htp]\centering
\scriptsize
\begin{tabular}{l|rrrr|rrrr}\toprule
\textbf{} &\multicolumn{4}{c|}{\textbf{Ranking}} &\multicolumn{4}{c}{\textbf{Linking}} \\\cmidrule{2-9}
&\textbf{Tiny} &\textbf{Small} &\textbf{Medium} &\textbf{Large} &\textbf{Tiny} &\textbf{Small} &\textbf{Medium} &\textbf{Large} \\\midrule
Contexts per Sample &10 &10 &10 &10 &10 &10 &10 &10 \\
Maximum Contexts &100 &100 &100 &100 &100 &100 &100 &100 \\
Masked &false &false &false &false &true &false &true &false \\
Batch Size &1 &1 &1 &1 &1 &1 &1 &1 \\
Subbatch Size &10 &10 &10 &10 &10 &10 &10 &10 \\
Learning Rate &1.00e-6 &1.00e-6 &1.00e-6 &1.00e-6 &1.00e-6 &1.00e-6 &1.00e-6 &1.00e-6 \\
Seed &3443300 &3250085 &4051229 &6248903 &9294686 &9773709 &9790717 &6248903 \\
\bottomrule
\end{tabular}
\caption{OWE (multi-context) - best model selected after grid search.}
\end{table}

\begin{table}[!htp]\centering
\scriptsize
\begin{tabular}{l|rrrr}\toprule
\textbf{} &\multicolumn{4}{c}{\textbf{ComplEx}} \\\cmidrule{2-5}
\textbf{} &\textbf{Tiny} &\textbf{Small} &\textbf{Medium} &\textbf{Large} \\\midrule
Embedding Dims. &300 &300 &300 &500 \\
Learning Rate &1 &0.6 &0.4 &0.6 \\
Regularizer Weight &0.3 &0.1 &0.1 &0.1 \\
Batch Size &64 &64 &64 &64 \\
Seed &1174584270 &2593575093 &2203942071 &2649116927 \\
\bottomrule
\end{tabular}
\caption{OWE reference embeddings - best model selected after grid search.}
\end{table}

\end{document}

%% file: glossary.tex
%
%

\newacronym{irt2}{IRT2}{inductive reasoning with text}

\newglossaryentry{todo}{name={[T]}, description=still to do}
\newglossaryentry{planned}{name={[P]}, description=planned or work in progress}
\newglossaryentry{tbd}{name={[D]}, description=to be discussed}

\newacronym[plural=KGs, firstplural=knowledge graphs (KG)]{kg}{KG}{knowledge graph}
\newacronym[plural=KMs, firstplural=knowledge models (KM)]{km}{KM}{knowledge model}
\newacronym[plural=KWs, firstplural=knowledge workers (KW)]{kw}{KW}{knowledge worker}
\newacronym[plural=KEs, firstplural=knowledge engineers (KE)]{ke}{KE}{knowledge engineer}

\newacronym{kgc}{KGC}{knowledge graph completion}
\newacronym{st}{KG}{knowledge graph}
\newacronym{ow-kgc}{OW-KGC}{open-world knowledge graph completion}
\newacronym{cw-kgc}{CW-KGC}{closed-world knowledge graph completion}
\newacronym{el}{EL}{entity linking}
\newacronym{ner}{NER}{named entity recognition}
\newacronym{re}{RE}{relation extraction}
\newacronym{kb}{KB}{knowledge base}

\newacronym{cbow}{CBOW}{Continuous Bag of Words}

\newacronym{cnn}{CNN}{Convolutional Neural Network}
\newacronym{loss-ce}{CE-Loss}{cross entropy loss}
\newacronym{loss-bce}{BCE-Loss}{binary cross entropy loss}
\newacronym{loss-mm}{MM-Loss}{max-margin loss}
\newacronym{s-ann}{ANN}{approximate nearest-neighbour}
\newacronym{e-mrr}{MRR}{mean reciprocal rank}
\newacronym{e-tp}{TP}{true positive}

\newacronym{ntn}{NTN}{Neural Tensor Networks}
\newacronym{dkrl}{DKRL}{Description-Embodied Knowledge Representation Learning}
\newacronym{mia}{MIA}{Multiple Interaction Attention}
\newacronym{irt}{IRT}{Inductive Reasoning with Text}

\newglossaryentry{bm25}{
  name={BM25},
  description={Vector space text retrieval model based on tf-idf}
}

\newglossaryentry{wikipedia}{
  name={Wikipedia},
  description={Open online encyclopedia}
}

\newglossaryentry{wikidata}{
  name={Wikidata},
  description={Community driven knowledge base}
}

\newglossaryentry{freebase}{
  name={Freebase},
  description={Community driven knowledge base incorporated by Google}
}

\newglossaryentry{dbpedia}{
  name={DBPedia},
  description={\gls{kb} continually extracted from \gls{wikipedia} infoboxes}
}

\newglossaryentry{fb15k}{
  name={FB15k},
  description={Closed-world Knowledge Graph Completion Dataset by \cited{bordes2013translating}}
}

\newglossaryentry{fb15k237}{
  name={FB15k237},
  description={\gls{cw-kgc} dataset introduced by \citet{toutanova-chen-2015-observer} based on \gls{fb15k} without inter-relation symmetry and redundancy}
}

\newglossaryentry{fb15k237-owe}{
  name={FB15k237-OWE},
  description={\gls{ow-kgc} dataset introduced by \citet{shah2019open} based on \gls{fb15k237} with freebase entity descriptions}
}

\newglossaryentry{codex}{
  name={CoDEx},
  description={\gls{cw-kgc} dataset introduced by \citet{safavi2020codex}}
}

\newglossaryentry{codex-m}{
  name={CoDEx-M},
  description={Mid-sized version of the \gls{cw-kgc} dataset introduced by \citet{safavi2020codex}}
}

\newglossaryentry{irt-fb}{
  name={IRT-FB},
  description={\gls{ow-kgc} dataset based on \gls{fb15k237}}
}

\newglossaryentry{irt-cde}{
  name={IRT-CDE},
  description={\gls{ow-kgc} dataset based on \gls{codex}}
}

\newglossaryentry{fb20k}{
  name={FB20k},
  description={Open-world \gls{kgc} dataset introduced by \citet{xie2016representation} based on \gls{fb15k}}
}

\newglossaryentry{dbpedia50k}{
  name={DBPedia50k},
  description={\gls{dbpedia} based open-world \gls{kgc} dataset introduced by \citet{shi2017open}}
}

\newglossaryentry{dbpedia500k}{
  name={DBPedia500k},
  description={\gls{dbpedia} based open-world \gls{kgc} dataset introduced by \citet{shi2017open}}
}

\newglossaryentry{distmult}{
  name={DistMult},
  description={Triple scorer by \citet{yang2014embedding}}
}

\newglossaryentry{conmask}{
  name={ConMask},
  description={Joint model for \gls{ow-kgc}}
}

\newglossaryentry{pykeen}{
  name={PyKEEN},
  description={Python Knowledge Embeddings}
}